%% file: main.tex
\documentclass[runningheads]{llncs}

\usepackage[final,year=2026]{eccv}
\usepackage{lmodern}
\usepackage{eccvabbrv}
\usepackage{graphicx}
\usepackage{booktabs}
\usepackage{array}
\usepackage{amsmath,amssymb}
\usepackage{algorithm}
\usepackage{algpseudocode}
\usepackage{placeins}
\usepackage[accsupp]{axessibility}
\usepackage{hyperref}

\graphicspath{{figure/}}
\newcommand{\method}{RefineRank}

\begin{document}

\title{RefineRank: Joint Box Refinement and Ranking for Surgical Spatio-Temporal Grounding}
\titlerunning{RefineRank for Surgical Grounding}
\author{Linzhe Jiang\inst{1}\thanks{Corresponding author.} \and
Jiayuan Huang\inst{2} \and
Changhao Zhang\inst{1} \and
Chunyang Jiang\inst{3} \and
Zhehua Mao\inst{1} \and
Mobarak I. Hoque\inst{4}}
\authorrunning{L. Jiang et al.}
\institute{UCL Hawkes Institute, University College London, London, UK\\
\email{linzhe.jiang.23@ucl.ac.uk} \and
Visual Understanding Research Group, Department of Informatics,\\
King's College London, London, UK \and
School of Medicine, Nankai University, Tianjin, China \and
Division of Informatics, Imaging and Data Sciences,\\
University of Manchester, Manchester, UK}
\maketitle

\ifdefined\publicresult
\begin{center}
\fbox{\parbox{0.91\textwidth}{\centering\small
Public result companion. This PDF is not the anonymous review submission.}}
\end{center}
\fi

\begin{abstract}
Surgical spatio-temporal grounding (STG) requires locating, at each video time
specified by a procedural question, the object that the question asks about.
Existing approaches face a trade-off: vision language models understand the
question context but produce imprecise coordinates, whereas open-set detectors
provide localized candidate boxes whose confidence does not reflect which box
answers the question. We introduce \method{}, which closes this gap at the
candidate-box level. A compact trainable module, RefineNet, combines the
language and regional features of a frozen medical vision language model with
the proposals of a frozen open-set detector: it predicts a bounded coordinate
correction and a quality score for every candidate box, and a fixed decoding
rule returns the original or refined box with the highest score. On the
MedVidBench Official Rankings (Verified), RefineRank records 0.421 STG mIoU,
the highest displayed STG score, while its global multi-metric rank is 11. In a
controlled evaluation on
separate training and evaluation videos, coordinate correction raises the
candidate~oracle upper bound from 0.6772 to 0.7302, and ranking the joint pool
of original and refined candidates by their RefineNet scores improves
STG mIoU from 0.2719 to 0.4534, whereas separately trained selectors over the
same pool reach at most 0.4186. These results show that a small box-level
module can reconcile question understanding with precise localization without
retraining either backbone.
Code is available at \url{https://github.com/linzhe001/RefineRank}.
\end{abstract}

\section{Introduction}
\label{sec:introduction}

Surgical spatio-temporal grounding (STG) answers procedural questions such as
``which instrument is the surgeon holding at 01:35?'' by locating the queried
object in the video. The question names a target, a temporal interval, and a
sampling step, which together determine an ordered set of requested video
times; the system must return one object box at each requested time. The target
may be a small instrument, an anatomical region, or a surgeon's hand, and it
can be occluded or visually similar to nearby objects. The task therefore
requires the spatial, temporal, and language interactions emphasized by prior
STG models \cite{yang2022tubedetr,jin2022stcat,gu2023cgstvg}, together with the
domain language and fine visual distinctions of surgical video
\cite{su2026medgrpo}.

Two model families address complementary halves of this problem. Medical vision
language models (MedVLMs), multimodal large language models adapted to medical
images and video, can interpret complex clinical questions, but semantic
understanding alone does not ensure accurate coordinates. Grounded multimodal
models such as Kosmos-2, Shikra, and Ferret add location tokens, coordinate
interfaces, regional representations, or dedicated grounding data to obtain
spatial outputs \cite{peng2023kosmos2,chen2023shikra,you2023ferret}. Open-set
detectors offer the complementary strength: given a text query, a detector such
as GroundingDINO \cite{liu2024groundingdino} returns a set of candidate boxes,
each with coordinates and a confidence score. That confidence measures how well
a box matches the detector query rather than how well it answers the complete
timestamped surgical question, so the box that best answers the question is
often not the highest-scoring one.

Combining the two families is nontrivial. Learning a dense alignment between
their feature spaces would require correspondences across different
tokenizations, dimensions, spatial grids, and pretraining objectives, a problem
that joint multimodal detectors address through grounded pretraining and
internal cross-modal fusion \cite{kamath2021mdetr,liu2024groundingdino}.
Learning a comparable alignment between two existing frozen backbones from
surgical STG data is harder still. Candidate boxes offer a compact alternative:
each proposed box already has coordinates, a detector score, and a matching
region in the MedVLM visual features, so the two models can be connected
through the boxes themselves rather than through their internal features.

RefineRank realizes this idea. It keeps a frozen MedVLM and a frozen
GroundingDINO detector and connects them only through the detector's candidate
boxes. The single trainable component, RefineNet, reads the MedVLM's language
and regional features at each candidate box and jointly learns two outputs: a
bounded correction that improves the box coordinates, and a quality score that
reflects how well the box answers the question. A fixed decoding rule with no
learned parameters returns the highest-scoring candidate from the joint pool of
original and refined boxes, so no separate selector module is needed. In this
way, question understanding improves detector grounding without learning a
dense alignment between the two feature spaces.

Our evaluation separates the two questions this design raises: whether
refinement creates better candidate boxes, and whether the learned scores
select better boxes. RefineNet's corrections raise the localization upper bound
of the candidate pool, its scores improve final selection over detector
confidence and over separately trained selectors, and a gap to the candidate
oracle remains.

We make three contributions. First, \textbf{RefineRank}, a complete grounding
pipeline whose compact trainable RefineNet module jointly learns scores and
corrections for detector candidates from MedVLM language and regional features
while both backbones remain frozen. Second, a \textbf{controlled evaluation}
that separates selection quality from localization potential by comparing the
final prediction with the best available box. Third, a \textbf{selector
analysis} showing that selectors trained only after RefineNet has modified the
boxes do not improve on the built-in decoding rule of RefineRank.
  
\section{Related Work}
\label{sec:related}

\paragraph{Spatio-temporal video grounding.}
TubeDETR directly predicts temporally localized tubes with transformer queries
\cite{yang2022tubedetr}. STCAT uses a single stage to maintain spatial and
temporal consistency \cite{jin2022stcat}. CG-STVG coordinates static and dynamic
vision language streams \cite{gu2023cgstvg}. These systems learn dense video
and language interactions. RefineRank instead uses RefineNet to improve and
rank detector boxes before applying a fixed rule that selects one
box at each requested time. The
requested timestamps are known, so the model does not predict an entire tube
directly from video tokens. The datasets and evaluators differ, so their
published scores are not compared numerically here.

\paragraph{Grounded vision language models.}
MDETR learns early multimodal fusion for detection conditioned on text
\cite{kamath2021mdetr}. GroundingDINO combines grounded pretraining with
tight language and vision fusion for open-set detection
\cite{liu2024groundingdino}. Multimodal language models approach the same
spatial problem through explicit grounding interfaces. Kosmos-2 represents
regions with location tokens \cite{peng2023kosmos2}. Shikra generates and
accepts coordinates in natural language \cite{chen2023shikra}. Ferret joins
coordinates with continuous region features \cite{you2023ferret}. These works
train grounding inside a single model. RefineRank instead forms a pipeline from
two frozen models and the compact RefineNet module after GroundingDINO has
proposed boxes that both models can refer to.

\paragraph{Medical video grounding.}
MedGRPO introduces MedVidBench and releases the
uAI-NEXUS-MedVLM-1.0a-7B-RL checkpoint \cite{su2026medgrpo}. RefineRank uses
this checkpoint as its frozen MedVLM, which supplies the detector query and
multimodal features, together with a separate frozen GroundingDINO detector. The
MedVLM + GroundingDINO baseline uses that query and ranks boxes by detector
confidence; it is not a trained fusion architecture. RefineNet is the only
trainable component that combines their outputs.

\paragraph{Localization quality and box refinement.}
Cascade R-CNN progressively raises proposal quality
\cite{cai2018cascadercnn}. IoU-Net predicts localization confidence
\cite{jiang2018iounet}. GFL and VarifocalNet align dense detector ranking
with localization quality \cite{li2020gfl,zhang2021varifocalnet}. The same
distinction between localization and ranking is relevant to surgical STG.
Unlike a detector head trained with its feature pyramid, RefineNet operates on
precomputed detector boxes and frozen multimodal features. Its two quality
outputs assign scores to the original and refined
versions of each proposal.

\section{Method}
\label{sec:method}

\begin{figure}[t]
  \centering
  \includegraphics[width=0.92\textwidth]{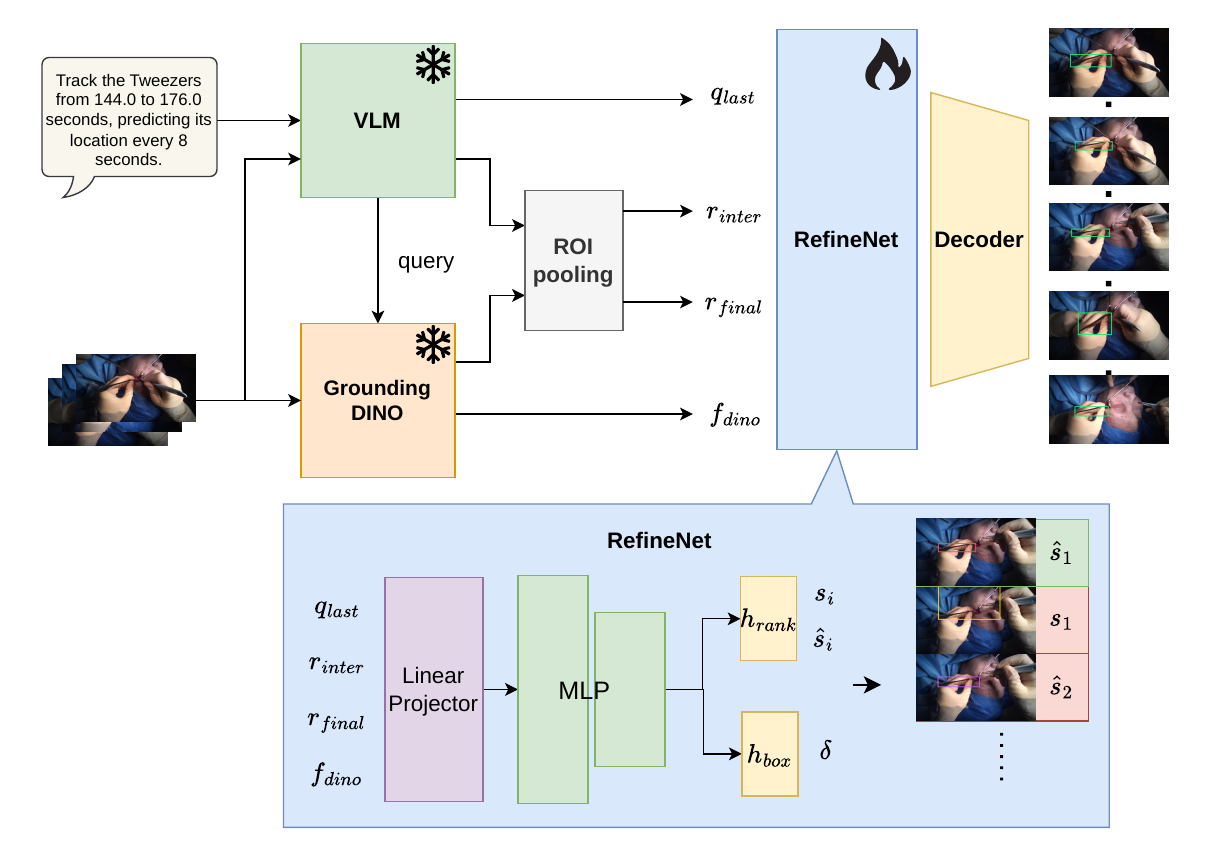}
  \caption{The complete \method{} pipeline. The
  frozen MedVLM reads the complete timestamped question and sampled frames. It
  supplies the detector query, $q_{\rm last}$, and two visual grids.
  GroundingDINO applies the query to the requested frames and provides original
  boxes and their metadata $f_{{\rm dino},i}$. ROI pooling produces
  $r_i^{\rm inter}$ and $r_i^{\rm final}$ for the trainable RefineNet module.
  Its ranking head $h_{\rm rank}$
  outputs logits $s_i$ and $\hat{s}_i$, while the box head $h_{\rm box}$ outputs
  the correction $\boldsymbol{\delta}_i$. The final pool contains original
  boxes scored by $s_i$ and refined boxes scored by $\hat{s}_i$. At each
  requested time, the fixed decoding rule returns the candidate with the highest
  corresponding score; it has no learned parameters and no separate selector
  module is used.}
  \label{fig:pipeline}
\end{figure}

The complete pipeline of RefineRank is shown in Figure~\ref{fig:pipeline}: the
frozen MedVLM, frozen GroundingDINO, the trainable RefineNet module, and a
fixed decoding rule that returns the candidate with the highest score. The
MedVLM supplies question and regional
features, GroundingDINO supplies box locations, and only RefineNet is
optimized. A GroundingDINO
detection before correction is called an \emph{original detector box}, or
simply an \emph{original box}. Its corrected version is called a
\emph{refined box}.

\subsection{Problem Setting}

The input question specifies a target, a temporal interval, and a sampling
step. These values determine an ordered set of requested times. The system must
return one axis-aligned box at each time, and STG mIoU averages box IoU over the
requested times. During training, $g$ denotes the target box. Each frame
contains several original boxes, and detector confidence does not necessarily
identify the box that answers the complete question.

RefineNet is the only trained component: GroundingDINO box generation, MedVLM
feature extraction, and final box selection remain fixed. This design also
bounds what the pipeline can recover. The complete system can rank and locally
correct an available original box, but it cannot locate a target if
GroundingDINO does not propose a box that covers it.

\subsection{Frozen Models and Original Detector Boxes}

The two frozen backbones provide complementary inputs: question understanding
and visual features from the MedVLM, and candidate boxes from GroundingDINO.
The frozen MedVLM receives the complete timestamped question and sampled video
frames. As illustrated in Figure~\ref{fig:pipeline}, it analyzes the question and
extracts target keywords as the detector query, exposes its final language state
$q_{\rm last}$ (the last-layer hidden state at the final prompt token), and
supplies intermediate and final visual grids. Frozen
GroundingDINO \cite{liu2024groundingdino} applies the detector query to each
requested RGB frame. Low detection thresholds favor recall. At most 12 raw
detections are retained. When a detection covers a broad region, fixed smaller
boxes derived from it are also added. Duplicate or invalid boxes are removed
before ranking.

The original box set is capped at 128 boxes per requested frame and is
constructed without target boxes. Each original box
$b_i=(x_{1i},y_{1i},x_{2i},y_{2i})$ is described by a metadata vector
$f_{{\rm dino},i}\in\mathbb R^{24}$ comprising 12 values and their 12
missing-value indicators. The values are the detector confidence; the upstream
temporal selection score; tube coverage, defined as the fraction of frames with
a detection; tube smoothness, defined as one minus the mean consecutive-box
center displacement normalized by the frame diagonal; the normalized time
offset $|t_i-t|/T$ and exact-match indicator
$\mathbf 1[|t_i-t|\le 10^{-6}]$; the four normalized box coordinates; and the
log area and log aspect ratio. Here, $t_i$ and $t$ denote the candidate and
requested times, and $T$ is the span of the sampled frames. Because
GroundingDINO is applied independently to each requested frame, every tube
contains one requested time: its temporal selection score and time offset are
zero, whereas its coverage, smoothness, and exact-match values are one. These
tube-level fields therefore preserve compatibility with multi-frame tubes but
do not vary across candidates in this pipeline. The values are standardized
with training-set statistics; an unavailable value is set to zero after
standardization and its indicator is set to one. Thus,
$f_{{\rm dino},i}$ describes detector output and box geometry rather than an
internal GroundingDINO feature.

The MedVLM is the uAI-NEXUS-MedVLM-1.0a-7B-RL checkpoint released by MedGRPO
\cite{su2026medgrpo}. For each requested frame, the same frozen model processes
that frame with the complete timestamped question. Its final language state is
$q_{\rm last}\in\mathbb{R}^{3584}$. The intermediate visual grid at block 23
and the final grid after visual token merging are retained. ROI pooling weighted
by area over the grid cells intersecting $b_i$ gives
$r_i^{\rm inter}\in\mathbb R^{1280}$ from block 23 and
$r_i^{\rm final}\in\mathbb R^{3584}$ from the final grid. Both features are
pooled at the original box coordinates. They are not recomputed after box
refinement. Weighting cells by their exact overlap with the box avoids rounding
a small box to a single nearest patch.

\subsection{Joint Ranking and Refinement}

RefineNet turns each candidate box into a joint score and correction from the
four inputs shown in Figure~\ref{fig:pipeline}: $q_{\rm last}$,
$r_i^{\rm inter}$, $r_i^{\rm final}$, and $f_{{\rm dino},i}$. The query and regional features are $\ell_2$-normalized,
and separate linear maps project all four inputs to 128 dimensions. The
projected regional and metadata features are added. An MLP with two layers combines
this box representation with the projected query and their elementwise
product. The ranking head $h_{\rm rank}$ outputs logits $s_i$ and $\hat{s}_i$
for the original and refined boxes. The box head $h_{\rm box}$ outputs
$\boldsymbol{\delta}_i=(\delta_{x,i},\delta_{y,i},\delta_{w,i},\delta_{h,i})$.
The scores are logits. A sigmoid is applied only when a probability is needed
for score calibration or candidate filtering. Since the sigmoid is monotonic,
it does not change their ranking.

A componentwise $\tanh$ bounds the center offsets $\delta_{x,i}$ and
$\delta_{y,i}$ to $[-0.5,0.5]$ and the logarithmic scale offsets
$\delta_{w,i}$ and $\delta_{h,i}$ to $[-\log 2,\log 2]$. Let the original box
$b_i$ have center $(c_{x,i},c_{y,i})$ and size $(w_i,h_i)$. RefineNet decodes
the corrected center and size as
\begin{equation}
\begin{aligned}
c'_{x,i} &= c_{x,i}+\delta_{x,i}w_i,
& c'_{y,i} &= c_{y,i}+\delta_{y,i}h_i,\\
w'_i &= w_i\exp(\delta_{w,i}),
& h'_i &= h_i\exp(\delta_{h,i}).
\end{aligned}
 \label{eq:box-decode}
\end{equation}
The result is converted to corner coordinates to obtain the refined box $b'_i$.
Coordinates are clipped to the normalized image range. Boxes with non-finite
coordinates or non-positive area are discarded. The same bounds are used when the
target $g$ is encoded relative to $b_i$ for regression.

Original and refined IoUs, $y_i=\operatorname{IoU}(b_i,g)$ and
$\hat y_i=\operatorname{IoU}(b'_i,g)$, supervise the two logits. Each score is
trained with the same $\textsc{RankingLoss}$,
\begin{equation}
\mathcal L_{\rm rank}(s,y)=
-\sum_{i\in\mathcal V} p_i\log q_i
+\lambda_{\rm cal}\,\ell_{\rm SL1}\big(\sigma(s),y\big),
 \label{eq:ranking-loss}
\end{equation}
where $\mathcal V$ is the set of valid (non-padded) boxes of one example,
$p=\operatorname{softmax}(y/\tau)$ with $\tau=0.1$ and
$q=\operatorname{softmax}(s)$ are distributions over $\mathcal V$, $\sigma$ is
the sigmoid, and $\ell_{\rm SL1}$ is the Smooth L1 loss
\cite{girshick2015fastrcnn} between the sigmoid scores and the IoU targets,
averaged over $\mathcal V$ and weighted by $\lambda_{\rm cal}=0.25$. The listwise term is averaged over examples with at
least one positive target IoU; examples with $\max_i y_i=0$ contribute only the
calibration term. The refined IoU target is detached from box decoding.
Box supervision uses
\begin{equation}
\mathcal L_{\rm box}=
\ell_{\rm SL1}\big(\boldsymbol{\delta}_{\mathcal I},\boldsymbol{\delta}_{\mathcal I}^{*}\big)
+\tfrac{1}{2}\,\ell_{\rm GIoU}\big(b'_{\mathcal I},g\big),
 \label{eq:box-loss}
\end{equation}
where $\mathcal I$ contains the eight valid original boxes with the highest
$y_i$, the subscript $\mathcal I$ denotes averaging over these boxes,
$\boldsymbol{\delta}_i^{*}$ encodes $g$ relative to $b_i$ under the
bounds of Eq.~\ref{eq:box-decode}, and
$\ell_{\rm GIoU}=1-\operatorname{GIoU}$ uses the generalized IoU
\cite{rezatofighi2019giou}. The total objective is the sum of the two
ranking losses and the box loss. Batches are padded to at most 128 boxes. A
Boolean mask removes padded boxes before softmax and from every loss.

\begin{algorithm}[t]
\caption{Training RefineNet}
\label{alg:training}
\begin{algorithmic}[1]
\Require Frozen features $q_{\rm last}$, $r^{\rm inter}$, $r^{\rm final}$,
$f_{\rm dino}$, original boxes $b$, targets $g$, and masks for valid boxes
\Require RefineNet parameters $\theta$ and optimizer $\mathcal O$
\Ensure Trained parameters $\theta$
\For{each minibatch}
  \State $(s,\hat{s},\boldsymbol{\delta}) \gets
  \operatorname{RefineNet}_{\theta}(q_{\rm last},r^{\rm inter},
  r^{\rm final},f_{\rm dino})$
  \State $b' \gets \Call{Decode}{b,\boldsymbol{\delta}}$
  \State $y \gets \operatorname{IoU}(b,g)$
  \State $\hat y \gets \Call{StopGrad}{\operatorname{IoU}(b',g)}$
  \State $\mathcal I \gets \Call{TopKValid}{y,8}$
  \State $\mathcal L_{\rm rank} \gets
  \Call{RankingLoss}{s,y}+\Call{RankingLoss}{\hat s,\hat y}$
  \State $\mathcal L_{\rm box} \gets
  \Call{BoxLoss}{b,b',\boldsymbol{\delta},g,\mathcal I}$
  \State $\mathcal L \gets \mathcal L_{\rm rank}+\mathcal L_{\rm box}$
  \State $\Call{ZeroGrad}{\mathcal O}$
  \State $\Call{Backward}{\mathcal L}$
  \State $\Call{ClipGradNorm}{\theta,1}$
  \State $\Call{Step}{\mathcal O}$
\EndFor
\end{algorithmic}
\end{algorithm}

Algorithm~\ref{alg:training} applies Eq.~\ref{eq:box-decode} in
\textsc{Decode}. \textsc{RankingLoss} is the objective of
Eq.~\ref{eq:ranking-loss}. \textsc{TopKValid} excludes padding and returns at
most eight boxes per example. \textsc{BoxLoss} is the objective of
Eq.~\ref{eq:box-loss} over these boxes.
Only $\theta$ enters the optimizer. The cached MedVLM and GroundingDINO outputs
receive no gradient.

\subsection{Candidate Pool and Final Selection}

At inference, RefineRank keeps both versions of every candidate so that an
accurate original box is never forced to move: the correction head produces
$(b'_i,\hat{s}_i)$ alongside each
original candidate and its score $(b_i,s_i)$. A fixed filter first keeps 16
original boxes, then greedily fills the remaining positions, up to 48
candidates, from the union of the original and refined boxes. Writing $s(c)$
and $b(c)$ for the RefineNet score and the box of candidate $c$, each step
adds the candidate that maximizes
\begin{equation}
U(c)=\lambda_{\rm q}\,\sigma\big(s(c)\big)+\lambda_{\rm d}\,d(c\mid\mathcal S),\qquad
d(c\mid\mathcal S)=1-\max_{c'\in\mathcal S}\operatorname{IoU}\big(b(c),b(c')\big),
 \label{eq:mmr}
\end{equation}
where $\mathcal S$ is the set of already selected candidates, $\sigma$ is
the sigmoid, and the quality and diversity weights are $\lambda_{\rm q}=0.7$
and $\lambda_{\rm d}=0.3$. The diversity term $d(c\mid\mathcal S)$ is one
minus the largest
IoU between $b(c)$ and any selected box, so it favors candidates that do not
duplicate an already selected location.

Final selection requires no learned module. Let $\mathcal C_t$ be the retained
candidates at requested time $t$. For
$c\in\mathcal C_t$, its RefineNet score is $s_i$ when $c=b_i$ and
$\hat{s}_i$ when $c=b'_i$, and the decoding rule simply returns
$\arg\max_{c\in\mathcal C_t}s(c)$. This rule over the joint pool of original
and refined boxes defines the primary result in
Sec.~\ref{sec:results}; no additional selector is part of the final RefineRank
prediction. The selector ablation replaces only these scores with separately
trained scoring rules while keeping the same candidate pool and argmax decoder.

\section{Experimental Protocol}
\label{sec:experiments}

\subsection{Training and Evaluation Data}

\begin{table}[t]
  \caption{Video-separated split of the controlled study. Each example is one
  requested timestamp with its sampled frame, question, and target box. The
  closed MedVidBench leaderboard \emph{test}$^{\dagger}$ split is not included.}
  \label{tab:split}
  \centering
  \small
  {\renewcommand{\arraystretch}{1.10}
  \begin{tabular*}{\textwidth}{@{\extracolsep{\fill}}lcccc@{}}
  \toprule
  & \multicolumn{2}{c}{\textbf{Videos}} & \multicolumn{2}{c}{\textbf{Examples}} \\
  \cmidrule(lr){2-3}\cmidrule(lr){4-5}
  \textbf{Dataset} & Train & Eval & Train & Eval \\
  \midrule
  CholecTrack20 & 7 & 2 & 807 & 212 \\
  \addlinespace[1.5pt]
  CoPESD & 5 & 2 & 134 & 118 \\
  \addlinespace[1.5pt]
  EgoSurgery & 11 & 3 & 1405 & 624 \\
  \midrule
  Total & 23 & 7 & 2{,}346 & 954 \\
  \bottomrule
  \end{tabular*}}
\end{table}

The daggered \emph{test} label denotes the closed MedVidBench leaderboard test
split: its ground-truth annotations are hidden and scores are returned only
through the benchmark server. The controlled study uses a fixed split of the STG portion of the MedVidU
training data,
spanning CholecTrack20 \cite{nwoye2025cholectrack20},
CoPESD \cite{wang2024copesd}, and EgoSurgery \cite{fujii2024egosurgerytool}. Each training example contains
one requested timestamp, its sampled frame, the complete question, and one
target box. The split covers 30 videos and 3{,}300 examples in total, divided
into 23 training videos with 2{,}346 examples and 7 evaluation videos with 954
examples (Table~\ref{tab:split}). Videos are assigned to either training or
evaluation, never both.
This assignment is shared by every controlled comparison. Results are reported
per dataset and by their simple mean. The leaderboard \emph{test}$^{\dagger}$
split is held out from this controlled split and is not used for training or
offline evaluation.

\subsection{Box and Feature Preparation}

Detector queries, original boxes, and MedVLM features are computed before
training RefineNet. The same frozen MedVLM supplies the
detector query and the features illustrated in Figure~\ref{fig:pipeline}. The
same precomputed detector boxes are used throughout the comparison. RefineNet
adds one refined box for every original box before applying the predefined
filtering step. Padding is excluded from evaluation.

Each MedVLM forward receives the requested RGB frame and the complete
timestamped question. The stored outputs are the final language state, the
visual grid after token merging, and visual blocks 7, 15, 23, and 31. Only
block 23 and the final
output are consumed by the main model. Blocks 7, 15, and 31 are retained for
the controlled layer comparison. Region features are pooled at the coordinates of the original
box. The refined box is not sent through the backbone again. GroundingDINO and
MedVLM remain in evaluation mode throughout feature extraction.

\subsection{Baselines and Reporting}

The \emph{direct MedVLM} baseline parses coordinates generated by the frozen
checkpoint. Invalid coordinate generations yield no box. The \emph{MedVLM +
GroundingDINO} baseline applies the MedVLM-derived query and selects from the
original boxes using detector confidence. \emph{RefineRank} constructs the
joint pool of original and refined boxes described in Sec.~\ref{sec:method},
ranks every retained candidate by its RefineNet score, and returns the box with
the highest score.
No additional selector is trained for this prediction. The selector ablation
uses the same pool and argmax decoder but replaces the RefineNet scores with
scores from separately trained selectors that are not components of RefineRank.

STG mIoU is computed at the annotated requested times for every method. The
\emph{candidate oracle} is also reported for diagnosis. It uses the target annotation to
choose the box with the greatest IoU at each time and is not a usable prediction method.
It measures the best result possible with the available boxes. For the MedVLM
+ GroundingDINO baseline, the candidate oracle considers only original boxes.
For RefineRank, it considers the union of original and refined boxes and
therefore measures the localization potential added by the correction head.

The public comparison uses the MedVidBench Official Rankings (Verified)
snapshot accessed on 15 August 2026.\footnote{\url{https://huggingface.co/spaces/UII-AI/MedVidBench-Leaderboard}}
The official leaderboard provides a global position obtained by averaging
per-metric ranks over ten metrics. Because this work concerns STG,
Table~\ref{tab:community} instead orders verified entries by STG mIoU and
reports the top five. RefineRank appears under the exact submission name
\emph{uAI-NEXUS-MedVLM-1.0a-7B-RL-STG\_final}.

\subsection{Training and Controls}

RefineNet has 1,251,334 trainable parameters and is trained for 40 epochs with
AdamW \cite{loshchilov2019adamw}. The learning rate is $3\times10^{-4}$, weight decay is $10^{-2}$, and
the gradient norm is clipped to 1. Batches are balanced across the three
datasets. All query and ROI tensors are precomputed and receive no gradients,
and the optimizer contains no GroundingDINO or MedVLM parameter.

The feature study uses a separate MLP that is trained for 10 epochs. It
operates on the joint pool of original and refined candidates produced by the
trained and frozen RefineNet: RefineNet supplies the refined boxes, but the
MLP re-scores every candidate independently and neither its input features nor
the selection rule use the RefineNet scores. Unlike the 40-epoch RefineNet, it
does not adjust boxes. Its purpose is to compare input
features, so it is not a variant of RefineNet. The training and evaluation
sets, the candidate pool, and final selection rule are
held fixed while the MLP input changes from metadata alone to combinations with
$q_{\rm last}$ and regional features from visual blocks 7, 15, 23, and 31.

For the selector ablation, RefineNet is first trained and then frozen. Its box
head generates the refined boxes, which are combined with the original boxes to
form the same fixed candidate pool used by RefineRank. ExtraTrees
\cite{geurts2006extratrees}, an MLP that scores each box independently from the
metadata, query, and block-23 regional features, and a
standard Transformer encoder \cite{vaswani2017attention} are then trained
separately on this fixed output. No selector receives the RefineNet scores as
an input feature. They do not update RefineNet and are not
components of RefineRank. Table~\ref{tab:rankers} includes RefineRank as the
reference row: no selector is trained, and its decoding rule ranks the
candidates directly by RefineNet's own scores. Each ablation selector
instead supplies replacement scores to the same argmax decoder.

\section{Results}
\label{sec:results}

\subsection{MedVidBench Official Benchmark}

\begin{table}[H]
  \caption{MedVidBench Official Rankings (Verified), accessed 15 August 2026.
  Entries are ordered by STG mIoU, and the top five are shown.}
  \label{tab:community}
  \centering
  \small
  {\renewcommand{\arraystretch}{1.10}\input{table/leaderboard_stg}}
\end{table}

Table~\ref{tab:community} lists the five highest STG mIoU results in the
official snapshot. RefineRank ranks first on this metric with 0.421, while its
global leaderboard rank is 11 because the global ordering aggregates ten
metrics.

\subsection{Does Refinement Create Better Candidate Boxes?}

\begin{table}[t]
  \caption{Controlled STG mIoU on separate training and evaluation videos.
  Dataset averages weight the three datasets equally.}
  \label{tab:controlled}
  \centering
  \small
  {\renewcommand{\arraystretch}{1.14}\input{table/controlled_results}}
\end{table}

Refinement expands what the candidate pool can contain. With the original
GroundingDINO boxes alone, the candidate oracle in
Table~\ref{tab:controlled}, the best available box chosen with annotations at
each requested time---reaches 0.6772. Adding the refined boxes produced by
RefineNet's correction head raises this upper bound to 0.7302. The correction
head therefore adds genuine localization potential beyond what the frozen
detector proposes on its own.

\subsection{Does the Learned Scoring Function Select Better Boxes?}

A well-localized box helps only if it is also selected. MedVLM-guided
GroundingDINO often proposes a useful box but does not assign it the highest
detector confidence: selecting by detector confidence gives an average of
0.2719, far below the 0.6772 candidate oracle over the same original boxes
(Table~\ref{tab:controlled}). Good candidates are thus often present but
poorly ranked for the complete surgical question.

RefineNet's learned scores close part of this gap. Ranking the joint pool of
original and refined boxes by their RefineNet scores reaches 0.4534, a gain of
0.1815 over the MedVLM + GroundingDINO baseline, with the largest improvements
on CholecTrack20 and CoPESD. RefineRank therefore improves both the candidate
boxes and their ranking, although a substantial gap to the 0.7302 candidate
oracle remains.

\subsection{Do Separately Trained Selectors Improve RefineRank?}

\begin{table}[t]
  \caption{Selector study on the same joint pool of original and refined
  candidates generated after RefineNet training. RefineNet is frozen for every row.
  RefineRank trains no additional selector: its decoding rule, which has no
  learned parameters, uses RefineNet's own scores. ExtraTrees, MLP, and the Transformer encoder are
  separately trained replacement scoring rules and are not RefineRank modules.
  None of them receives the RefineNet scores as an input feature. The MLP
  combines the metadata, query, and block-23 regional features; it is the same
  configuration as the block-23 row of Table~\ref{tab:layers} and therefore
  reports the same values.}
  \label{tab:rankers}
  \centering
  \small
  {\renewcommand{\arraystretch}{1.10}\input{table/ranker_controls}}
\end{table}

Better refined boxes do not make selection automatic. RefineRank's decoding
rule uses RefineNet's own scores directly and reaches 0.4534 on the
joint pool. After RefineNet training is complete, the model is frozen and the
strongest separately trained selector ablation is the MLP at 0.4186;
ExtraTrees and the Transformer encoder also remain below RefineRank. Because all
four rows use the same RefineNet-generated candidate pool and argmax decoder,
the comparison isolates whether replacing RefineNet's own scores with an
additional learned scoring rule improves selection. In this study, none does.
The result indicates that RefineNet already learns the
strongest evaluated box-quality signal, while the remaining candidate oracle
gap shows that its scores still do not always promote the best available box.

\section{Ablations and Qualitative Analysis}
\label{sec:analysis}

\subsection{Input Feature Study with a Separate Box Scorer}

\begin{table}[t]
  \caption{Input feature study with a separately trained box scoring MLP. The
  candidate pool is the joint pool of original and refined boxes generated by the
  trained and frozen RefineNet, and every row re-scores all of its candidates
  after 10 epochs of MLP training. The RefineNet scores are used neither as
  input features nor at selection time.}
  \label{tab:layers}
  \centering
  \small
  {\renewcommand{\arraystretch}{1.03}\input{table/layer_ablation}}
\end{table}

A separate 10-epoch MLP provides the input study reported in
Table~\ref{tab:layers}. It is not an ablation of $h_{\rm rank}$ in the trained
RefineNet module. RefineNet is trained and frozen first, and its correction
head generates the refined half of the candidate pool; the MLP then re-scores
every candidate of this joint pool and changes one feature
input at a time. Ranking uses only the MLP's own scores: the RefineNet scores
$s_i$ and $\hat s_i$ are neither MLP inputs nor selection signals. Its values
need not match the RefineRank results in Table~\ref{tab:controlled}.
The block-23 row is the same configuration as the MLP row of
Table~\ref{tab:rankers}, so the two tables report identical values for it.
The $f_{\rm dino}$ row uses the 24 dimensional metadata
vector and is not the raw GroundingDINO confidence baseline. Within this
diagnostic model, adding $q_{\rm last}$ raises the average from 0.2767 to
0.4044. Intermediate visual features provide a smaller additional gain.
Block 23 has the highest equally weighted average, but blocks 7, 15, and 23 remain
closely grouped and their order varies by dataset. Block 23 is used
as the aggregate choice, without treating one depth as universally superior.

\subsection{Spatial Response Across Depth}

\begin{figure}[H]
  \centering
  \includegraphics[width=0.9\textwidth]{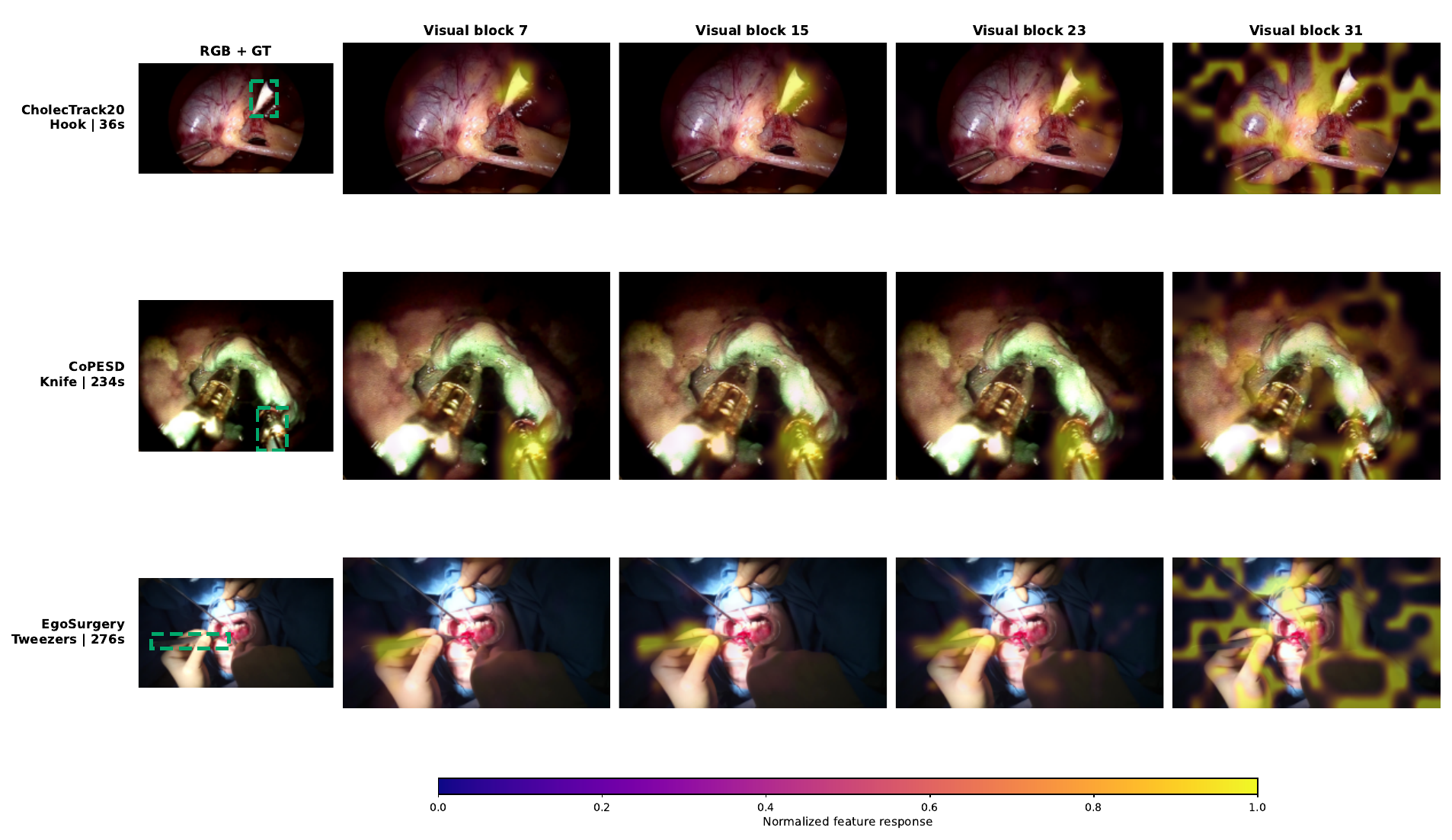}
  \caption{Spatial responses across frozen MedVLM visual blocks 7, 15, 23,
  and 31. For this visualization only, target region features are pooled after
  the multimodal forward and compared with every spatial token. Each map is
  normalized independently and is not an attention map or a causal explanation.}
  \label{fig:attribution}
\end{figure}

The response maps in Figure~\ref{fig:attribution} show localized evidence around
the displayed tools in intermediate blocks, whereas block 31 is less spatially
differentiated in these examples. This visual pattern is consistent with the
aggregate ablation but does not explain it causally. The target is used only to
construct this response map, never as a RefineNet input, and each map
is normalized independently. Neither color intensity nor spatial spread should
be interpreted as calibrated confidence or attention.

\FloatBarrier
\subsection{Examples from the Same Frame}

\begin{figure}[H]
  \centering
  \includegraphics[width=\textwidth]{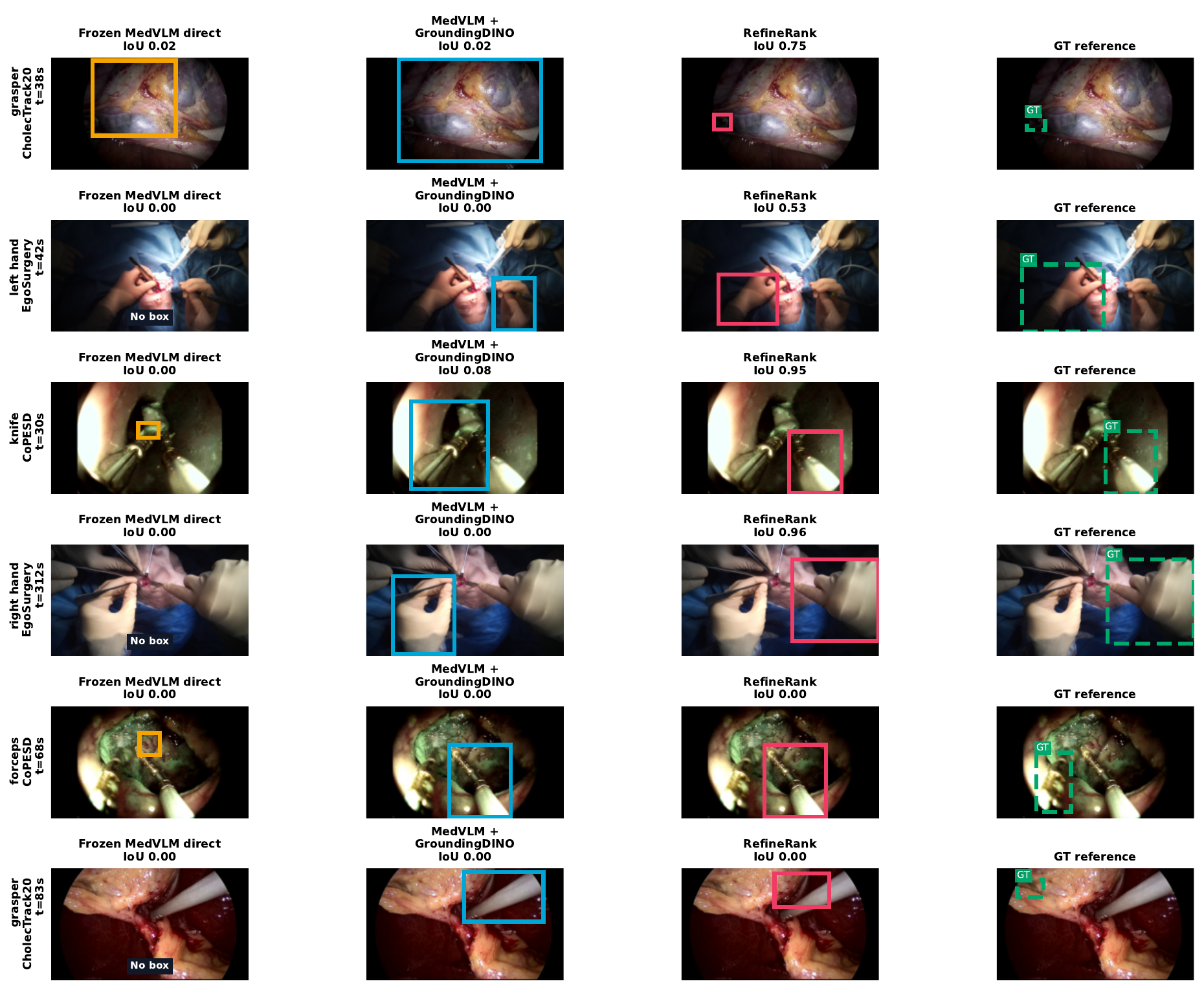}
  \caption{Six representative evaluation examples. Columns show direct
  MedVLM, MedVLM + GroundingDINO, RefineRank, and target boxes. The bottom two
  rows show shared failures. A displayed RefineRank prediction may therefore be
  either an original box or its refined counterpart, whichever received the
  higher score.}
  \label{fig:cases}
\end{figure}

The examples in Figure~\ref{fig:cases} illustrate how selecting an original box
and refining its position are complementary. A refined box improves one hand
localization, while several examples are solved by assigning a higher score to
a better original detector box without moving it. In the two shared failures,
the available proposals do not support a correct selection. These examples
illustrate the correction and scoring roles but do not estimate how frequently
either behavior occurs.

\subsection{Limitations}

The controlled analysis uses one fixed video-separated split so that all
methods share the same data assignment. Evaluation on additional splits would
further establish stability and is left to future work. The leaderboard result
is taken from the MedVidBench Official Rankings (Verified) snapshot accessed on
15 August 2026. No learned fusion baseline between MedVLM and GroundingDINO was trained or
evaluated.
Consequently, the controlled comparison shows an improvement over direct
MedVLM coordinates and the MedVLM + GroundingDINO baseline, but it does not
establish that fusion based on boxes is superior to other learned fusion
designs.

Regional features are pooled only at the original detector coordinates.
A refined box is not encoded again, so its score cannot use visual content newly
included or excluded by the correction. RefineRank also cannot recover a target
when GroundingDINO does not propose a box that covers it. Candidate oracle values use
target annotations and only show an upper bound, while the spatial response
maps are descriptive rather than explanations of individual scores.

\section{Conclusion}
\label{sec:conclusion}

RefineRank is a complete pipeline that combines the question understanding of a
frozen MedVLM, the localized candidate boxes of a frozen GroundingDINO
detector, and the compact trainable RefineNet module.
A fixed decoding rule then ranks the joint pool of original and refined
candidates by their RefineNet scores and returns the box with the highest
score, with no additional selector module. This built-in rule is the strongest
evaluated selection policy, and selectors trained separately on the fixed
RefineNet outputs do not improve it. The gap to the \mbox{candidate oracle}
therefore reflects both localization and ranking errors. Further evaluation
should cover other video sets and learned fusion baselines.
\FloatBarrier
\bibliographystyle{splncs04}
\bibliography{references}

\end{document}

%% file: table/leaderboard_stg.tex
\begin{tabular*}{\textwidth}{@{\extracolsep{\fill}}>{\raggedright\arraybackslash}p{0.78\textwidth}r@{}}
\toprule
\textbf{MedVidBench Leaderboard} & \textbf{STG mIoU} \\
\midrule
uAI-NEXUS-MedVLM-1.0a-7B-RL-STG\_final \textbf{(ours)} & \textbf{0.421} \\
\addlinespace[1.5pt]
uAI-NEXUS-MedVLM-1.0a-7B-RL~\cite{su2026medgrpo} & 0.202 \\
\addlinespace[1.5pt]
uAI-NEXUS-MedVLM-1.0c-4B-SFT & 0.190 \\
\addlinespace[1.5pt]
uAI-NEXUS-MedVLM-1.0a-7B-SFT & 0.177 \\
\addlinespace[1.5pt]
uAI-NEXUS-MedVLM-1.0b-4B-RL & 0.176 \\
\bottomrule
\end{tabular*}

%% file: table/controlled_results.tex
\begin{tabular*}{\textwidth}{@{\extracolsep{\fill}}>{\raggedright\arraybackslash}p{0.34\textwidth}rrrrr@{}}
\toprule
\textbf{Method} & \multicolumn{3}{c}{\textbf{Dataset STG mIoU}} & \textbf{Selected avg.} & \textbf{Oracle avg.} \\
\cmidrule(lr){2-4}
 & Cholec & CoPESD & Ego & & \\
\midrule
MedVLM~\cite{su2026medgrpo} & 0.0929 & 0.0170 & 0.0190 & 0.0429 & 0.0429 \\
\addlinespace[2pt]
MedVLM + GroundingDINO~\cite{su2026medgrpo,liu2024groundingdino} & 0.1350 & 0.3371 & 0.3435 & 0.2719 & 0.6772 \\
\addlinespace[2pt]
RefineRank (ours) & \textbf{0.3960} & \textbf{0.5972} & \textbf{0.3671} & \textbf{0.4534} & \textbf{0.7302} \\
\bottomrule
\end{tabular*}

%% file: table/ranker_controls.tex
\begin{tabular*}{\textwidth}{@{\extracolsep{\fill}}>{\raggedright\arraybackslash}p{0.42\textwidth}rrrr@{}}
\toprule
\textbf{Candidate scoring rule} & \multicolumn{3}{c}{\textbf{Dataset STG mIoU}} & \textbf{Dataset avg.} \\
\cmidrule(lr){2-4}
 & Cholec & CoPESD & Ego & \\
\midrule
ExtraTrees~\cite{geurts2006extratrees} & 0.2842 & 0.4243 & 0.2781 & 0.3289 \\
\addlinespace[1.5pt]
MLP & 0.3373 & 0.5767 & 0.3417 & 0.4186 \\
\addlinespace[1.5pt]
Transformer encoder~\cite{vaswani2017attention} & 0.3447 & 0.5415 & 0.3261 & 0.4041 \\
\addlinespace[1.5pt]
Native RefineNet scores (RefineRank) & \textbf{0.3960} & \textbf{0.5972} & \textbf{0.3671} & \textbf{0.4534} \\
\bottomrule
\end{tabular*}

%% file: table/layer_ablation.tex
\begin{tabular*}{\textwidth}{@{\extracolsep{\fill}}>{\raggedright\arraybackslash}p{0.43\textwidth}rrrr@{}}
\toprule
\textbf{Candidate MLP input} & \multicolumn{3}{c}{\textbf{Dataset STG mIoU}} & \textbf{Dataset avg.} \\
\cmidrule(lr){2-4}
 & Cholec & CoPESD & Ego & mean \\
\midrule
$f_{dino}$ & 0.1512 & 0.3578 & 0.3213 & 0.2767 \\
\addlinespace[1.5pt]
$f_{dino}$ + $q_{last}$ & 0.2528 & \textbf{0.6069} & 0.3536 & 0.4044 \\
\addlinespace[1.5pt]
$f_{dino}$ + $q_{last}$ + visual block 7 & \textbf{0.3779} & 0.5488 & 0.3260 & 0.4176 \\
\addlinespace[1.5pt]
$f_{dino}$ + $q_{last}$ + visual block 15 & 0.3537 & 0.5411 & \textbf{0.3558} & 0.4169 \\
\addlinespace[1.5pt]
$f_{dino}$ + $q_{last}$ + visual block 23 & 0.3373 & 0.5767 & 0.3417 & \textbf{0.4186} \\
\addlinespace[1.5pt]
$f_{dino}$ + $q_{last}$ + visual block 31 & 0.2622 & 0.6040 & 0.3213 & 0.3959 \\
\bottomrule
\end{tabular*}